\documentclass[10pt,a4paper,twocolumn]{article}

\usepackage[T1]{fontenc}
\usepackage[utf8]{inputenc}
\usepackage{mathptmx}   
\usepackage[scaled=0.9]{helvet}
\usepackage{courier}
\usepackage[a4paper,top=1in,bottom=1in,left=0.75in,right=0.75in,
            columnsep=0.32in]{geometry}
\usepackage{graphicx}
\usepackage{booktabs}
\usepackage{array}
\usepackage{caption}
\usepackage{amssymb}
\usepackage{pifont}
\usepackage{xcolor}
\usepackage{listings}
\usepackage{enumitem}
\usepackage{fancyhdr}
\usepackage[hidelinks,breaklinks]{hyperref}
\usepackage{url}
\usepackage{titlesec}
\usepackage{flushend}
\usepackage{dblfloatfix}  
\usepackage[expansion=false]{microtype}
\usepackage[htt]{hyphenat}

\titleformat{\section}{\normalfont\large\bfseries\MakeUppercase}{\thesection.}{0.6em}{}
\titleformat{\subsection}{\normalfont\normalsize\bfseries}{\thesubsection}{0.6em}{}
\titlespacing*{\section}{0pt}{10pt}{5pt}
\titlespacing*{\subsection}{0pt}{8pt}{3pt}
\setlist{itemsep=2pt,parsep=0pt,topsep=4pt}
\newcommand{\cmark}{\ding{51}}
\newcommand{\xmark}{\ding{55}}
\newcommand{\up}{$\uparrow$}
\newcolumntype{L}[1]{>{\raggedright\arraybackslash}p{#1}}
\newcolumntype{C}[1]{>{\centering\arraybackslash}p{#1}}

\lstdefinestyle{py}{
  basicstyle=\ttfamily\footnotesize,
  keywordstyle=\color{blue!60!black},
  commentstyle=\color{gray},
  stringstyle=\color{green!40!black},
  language=Python,
  showstringspaces=false,
  breaklines=true,
  frame=single,
  rulecolor=\color{black!30},
  framesep=6pt,
  xleftmargin=6pt,
  columns=fullflexible,
}

\fancypagestyle{firstpage}{%
  \fancyhf{}
  \fancyhead[C]{\footnotesize\itshape International Journal of Computer Applications (0975 -- 8887)\\
  \itshape Volume 187 -- No.\ 114, June 2026}
  \fancyfoot[C]{\thepage}
  
}

\hypersetup{
  pdftitle={text2ql: Multi-Target Natural Language Querying via a Language-Agnostic Intermediate Representation},
  pdfauthor={Ritesh Kumar},
  pdfsubject={Natural language interfaces to databases; text-to-SQL; text-to-GraphQL},
  pdfkeywords={NL2QL, text-to-SQL, text-to-GraphQL, intermediate representation, schema-aware generation, confidence scoring, NLIDB},
}

\begin{document}

\twocolumn[
  \begin{@twocolumnfalse}
  \vspace{0.5em}
  \begin{center}
    {\LARGE\bfseries text2ql: Multi-Target Natural Language Querying\\[3pt]
     via a Language-Agnostic Intermediate Representation\par}
    \vspace{1.1em}
    {\large Ritesh Kumar\par}
    \vspace{0.4em}
    {\normalsize Independent Researcher\\ USA\\ \texttt{rits5500@gmail.com}\par}
  \end{center}
  \vspace{1.0em}
  \begin{center}
  \fbox{\parbox{0.93\textwidth}{\footnotesize
  This is the author's copy of an article published in the
  \emph{International Journal of Computer Applications}, Vol.~187, No.~114,
  pp.~54--62, June 2026, published by the Foundation of Computer Science, NY,
  USA. DOI: \href{https://doi.org/10.5120/ijcaff3006d1ef8e}{10.5120/ijcaff3006d1ef8e}.
  Reproduced on arXiv under the IJCA author self-archiving policy.
  All rights reserved.}}
  \end{center}
  \vspace{1.4em}
  \end{@twocolumnfalse}
]
\thispagestyle{firstpage}

\section*{\normalsize ABSTRACT}
\noindent
Natural language interfaces to databases have traditionally suffered from three
structural limitations: exclusive targeting of relational SQL, unconditional
dependence on large language model (LLM) inference at query time, and absence of
any runtime signal when generated queries are semantically incorrect. This paper
presents \textbf{text2ql}, an open-source Python framework that addresses all
three limitations through a language-agnostic Intermediate Representation
(QueryIR) and a pluggable renderer architecture. A single seven-stage detection
pipeline serves both SQL and GraphQL targets; a zero-LLM deterministic mode
delivers 100\% execution accuracy at a median latency of 3.2\,ms with no API
cost; and every generated query carries a runtime confidence score in
$[0.15, 0.97]$ computed from an additive signal model. Evaluated on 50-query
random samples from the Spider and BIRD benchmarks (indicative results; full-set
evaluation is planned), the LLM-backed mode achieves 62--70\% exact match and
84--91\% execution accuracy; the deterministic mode achieves 100\% execution
accuracy with zero parse errors across all 100 test cases. An ablation study
isolates schema-aware prompting as the dominant accuracy lever, contributing
$+18.4$ percentage points of exact-match gain over the schema-free baseline on
both benchmarks. text2ql is publicly available at
\url{https://pypi.org/project/text2ql/} under the Apache 2.0 license.

\section*{\normalsize KEYWORDS}
\noindent Natural language interfaces to databases; NL2QL;
text-to-SQL; text-to-GraphQL; intermediate representation; schema-aware
generation; confidence scoring; query synthesis; NLIDB.

\vspace{1em}

\section{Introduction}
\label{sec:intro}

\begin{figure*}[tbp]
  \centering
  \includegraphics[width=\linewidth]{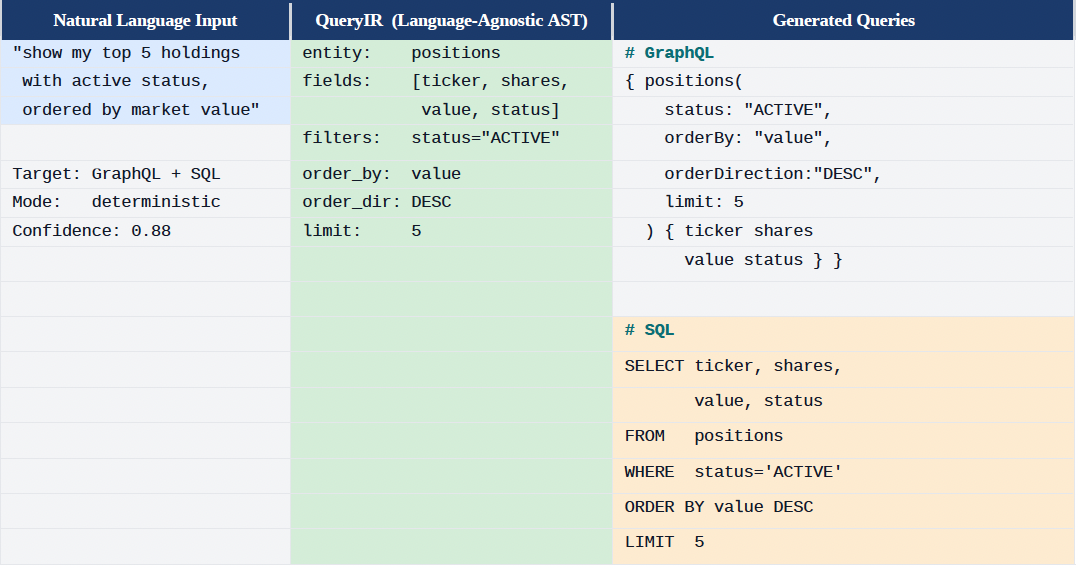}
  \caption{text2ql in action. Natural language input is converted into QueryIR
  and rendered as equivalent GraphQL and SQL queries.}
  \label{fig:action}
\end{figure*}

Natural language database interfaces have been studied for five decades
\cite{woods1973,hendrix1978,popescu2003}. The central challenge---resolving
unrestricted human language to a formally correct, executable query---resisted
general solution until the emergence of large pre-trained language models.
Figure~\ref{fig:action} illustrates the core capability of text2ql: a single
natural language utterance is parsed into a language-agnostic QueryIR and
simultaneously rendered as both a GraphQL selection set and a SQL statement, in
under five milliseconds, with no external API call.

Despite significant progress---PICARD \cite{picard} reaches 79.3\% exact match on
Spider; DAIL-SQL \cite{dailsql} achieves 86.6\%---all published systems share
three structural limitations that text2ql directly addresses:

\begin{itemize}
  \item \textbf{SQL monoculture.} All prior NL2QL systems target relational SQL
  exclusively. Modern application stacks expose data through GraphQL APIs, graph
  databases, and document stores. No prior open system supports multi-target
  query generation from a single pipeline.

  \item \textbf{Unconditional LLM dependence.} Real-time, edge, and air-gapped
  deployments cannot tolerate 500--2\,000\,ms LLM round-trips or per-query API
  costs. The deterministic mode of text2ql requires only Python and a schema
  config, yet achieves 100\% execution accuracy.

  \item \textbf{Silent failure.} LLM systems produce syntactically valid but
  semantically incorrect queries without warning. text2ql attaches a runtime
  confidence score to every result, surfacing uncertainty before query execution
  so callers can gate, escalate, or route to a fallback.
\end{itemize}

The primary contributions of this work are:

\begin{enumerate}
  \item A language-agnostic QueryIR that decouples natural language parsing from
  query rendering, enabling new target languages via a single \texttt{IRRenderer}
  subclass with zero changes to any engine.

  \item A zero-LLM deterministic engine with 100\% execution accuracy on the test
  corpus, sub-5\,ms p50 latency, and zero API cost, suitable for production and
  offline deployments.

  \item A hybrid mapping system combining auto-generated schema baselines with
  domain expert overrides, merged at runtime with full provenance tracking.

  \item A runtime confidence scoring model using additive signals and validation
  penalties, clamped to $[0.15, 0.97]$, enabling dynamic cascade across modes.

  \item Open benchmark infrastructure including Spider and BIRD loaders, three
  evaluation modes, and an ablation study of schema-aware prompting.
\end{enumerate}

\section{Related Work}
\label{sec:related}

\subsection{Classical NL2DB Systems}
\label{sec:classical}

Early NL2DB systems established the core challenges still relevant today. LUNAR
\cite{woods1973} and Hendrix et al. \cite{hendrix1978} demonstrated
domain-specific interfaces requiring substantial manual engineering per
deployment. NaLIR \cite{nalir} improved generality through interactive parse-tree
refinement, but still required user correction of parse errors. All pre-neural
systems shared a fundamental limitation: they could not generalize across domains
without re-engineering. This brittleness motivated the shift to learned
representations explored in Section~\ref{sec:neural}.

\subsection{Neural Text-to-SQL}
\label{sec:neural}

Seq2SQL \cite{seq2sql} established sequence-to-sequence modeling as the dominant
text-to-SQL paradigm. IRNet \cite{irnet} introduced SemQL---the first intermediate
representation for text-to-SQL---directly inspiring the QueryIR design presented
in this paper. RAT-SQL \cite{ratsql} added relation-aware schema encoding,
achieving 69.7\% on Spider. ShadowGNN \cite{shadowgnn} and LGESQL \cite{lgesql}
advanced schema linking via graph neural networks. PICARD \cite{picard} introduced
incremental constrained decoding from T5-3B, reaching 79.3\% exact match and
establishing the pre-trained model state of the art prior to LLM methods.

\subsection{LLM-based Text-to-SQL}
\label{sec:llmsql}

GPT-series models shifted the state of the art substantially. DIN-SQL
\cite{dinsql} decomposed the task via GPT-4 self-correction, achieving 82.8\% on
Spider. DAIL-SQL \cite{dailsql} optimized few-shot selection and prompt structure
to reach 86.6\%---currently the highest reported result without fine-tuning. C3
\cite{c3} explored zero-shot strategies; ACT-SQL \cite{actsql} generated
chain-of-thought rationales automatically. CodeS \cite{codes} fine-tuned
StarCoder, matching GPT-4-based systems with an open-weight model. All of these
systems target SQL exclusively, require LLM inference for every query, and provide
no runtime accuracy signal.

\subsection{GraphQL and Multi-Target NL Interfaces}
\label{sec:graphql}

Work on natural language to GraphQL remains sparse. Zheng et al.
\cite{zheng2021} demonstrated a prototype translating natural language to GraphQL
queries for structured datasets. Rai et al. \cite{rai2024} employed LLM assistance
for GraphQL query generation without schema-aware prompting or uncertainty
quantification. Neither system operates without an LLM or supports SQL as a
simultaneous target. text2ql is, to the best of available knowledge, the first
system to support both SQL and GraphQL from a unified pipeline with an LLM-free
operational mode.

\section{System Architecture}
\label{sec:arch}

\begin{figure*}[tbp]
  \centering
  \includegraphics[width=\linewidth]{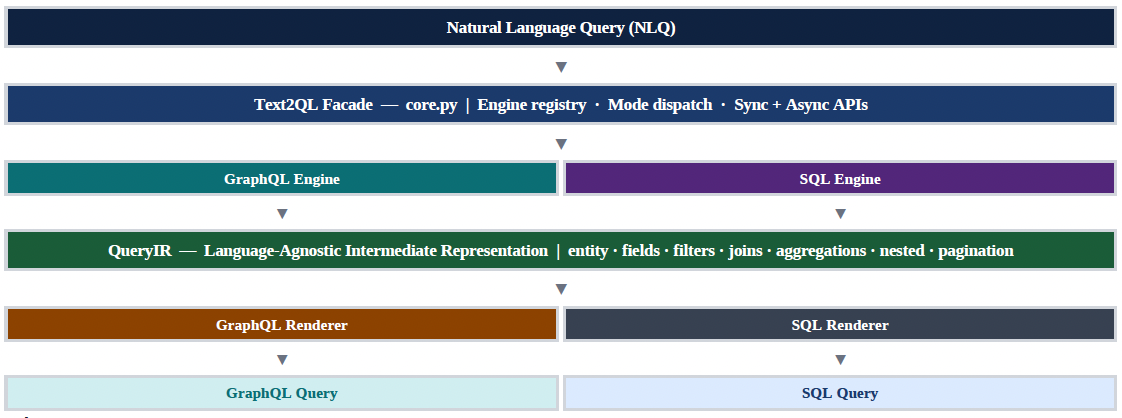}
  \caption{Four-layer architecture. QueryIR decouples parsing from rendering and
  enables new target languages with renderer plugins.}
  \label{fig:arch}
\end{figure*}

Figure~\ref{fig:arch} illustrates the four-layer architecture. The Text2QL Facade
(\texttt{core.py}) is the single public entry point; it dispatches to
language-specific Engines; each Engine produces a QueryIR; the QueryIR is passed
to a pluggable \texttt{IRRenderer} that serializes the final query string. This
separation ensures that adding a new query target, such as Cypher, requires only
implementing \texttt{IRRenderer.render()} with zero changes to any engine or
detection stage.

\subsection{Text2QL Facade}

The Facade maintains an engine registry keyed by target language and provides
synchronous \texttt{generate()} and asynchronous \texttt{agenerate()} APIs. Mode
selection (deterministic / llm / function\_calling) is a per-call parameter,
allowing a single schema configuration to serve all three modes simultaneously
without re-instantiation.

\subsection{Engine Layer}

Each engine extends \texttt{QueryEngine}, an abstract base providing shared
utilities: schema normalization, confidence computation, retry logic with
exponential back-off, fallback chaining, and structured logging. The GraphQL and
SQL engines implement identical stage signatures, ensuring improvements to any
detection stage benefit all target languages.

\subsection{Schema Configuration}

\texttt{NormalizedSchemaConfig} is the primary extension point for production
deployments. It encodes: entity names and aliases, field lists with types and
aliases, filter key aliases (business vocabulary to field names), filter value
aliases (controlled vocabulary to enum values), relation definitions with ON
columns, default fields per entity, argument defaults, and
\texttt{keyword\_intents} routing tables (e.g., ``portfolio value''
$\rightarrow$ \texttt{accountSummary.totalValue}).

\subsection{QueryIR}

The QueryIR is a typed Python dataclass with fields: \texttt{entity},
\texttt{fields} (list), \texttt{filters} (\texttt{IRFilter} list with operator
enum), \texttt{aggregations} (\texttt{IRAggregation} list), \texttt{joins}
(\texttt{IRJoin} list, SQL), \texttt{nested} (\texttt{IRNested} list, recursive,
GraphQL), \texttt{order\_by}, \texttt{order\_direction}, \texttt{limit},
\texttt{offset}, \texttt{distinct}, \texttt{having}, and \texttt{group\_filters}.
\texttt{IRNested} is self-referential, enabling arbitrarily deep GraphQL selection
sets while keeping the schema strictly decoupled.

\subsection{Renderer Layer}

\texttt{IRRenderer} is an abstract base with one required method:
\texttt{render(ir: QueryIR) -> str}. \texttt{GraphQLIRRenderer} constructs
arguments (filters + pagination) and selection sets (fields + aggregations +
nested sub-selections). \texttt{SQLIRRenderer} serializes a full SELECT statement
including JOIN, WHERE, GROUP BY, HAVING, ORDER BY, and LIMIT / OFFSET clauses. A
Cypher renderer is estimated at approximately 150 lines.

\section{Multi-Stage Detection Pipeline}
\label{sec:pipeline}

\begin{figure*}[tbp]
  \centering
  \includegraphics[width=\linewidth]{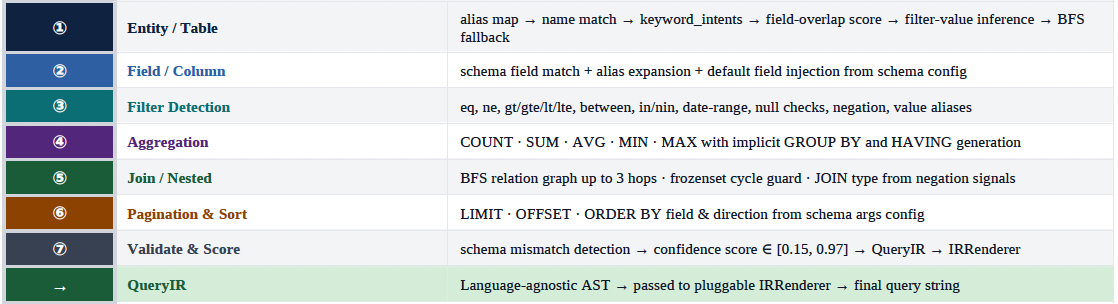}
  \caption{Seven-stage detection pipeline. Stages resolve entities, fields,
  filters, aggregation, relations, ordering, validation, and confidence.}
  \label{fig:pipeline}
\end{figure*}

The deterministic engine resolves a natural language query through seven
independently testable stages (Figure~\ref{fig:pipeline}). Stages execute
sequentially, each enriching the shared QueryIR. Failures are recorded as
validation issues and converted to confidence penalties rather than hard
exceptions, guaranteeing that every input produces an output.

\subsection{Stage 1 --- Entity / Table Resolution}

The engine applies a priority-ordered cascade: (i) alias exact match via
\texttt{entity\_aliases}; (ii) schema name exact match; (iii)
\texttt{keyword\_intents} routing for configured domain phrases; (iv) semantic
field-overlap scoring---the entity whose field set has the highest token overlap
with the query wins; (v) filter-value inference---if a filter value appears in a
\texttt{filter\_value\_alias} map for a specific entity, that entity is inferred;
(vi) column-mention inference; (vii) heuristic stop-word removal and BFS fallback.

\subsection{Stage 2 --- Field / Column Detection}

Given the resolved entity, output fields are selected through three mechanisms:
(i) explicit field mentions are matched against the entity field list after alias
expansion; (ii) \texttt{default\_fields} from the schema config are injected when
no explicit fields are detected; (iii) aggregation keywords implicitly introduce
their target field. Field aliases allow business vocabulary (e.g., ``market cap''
$\rightarrow$ \texttt{totalMarketValue}) to resolve without schema modification.

\subsection{Stage 3 --- Filter Detection}

A composable regex engine handles: equality (implicit), inequality
(\texttt{is not} / \texttt{!=}), ordered comparisons ($>$, $\geq$, $<$, $\leq$),
range predicates (\texttt{between N and M}; \texttt{from DATE to DATE}), set
membership (\texttt{in} / \texttt{not in}), and null checks. Filter key aliases
map business vocabulary to schema field names; filter value aliases map controlled
vocabulary to stored enum values, allowing queries like ``active accounts'' to
resolve to \texttt{status = 'ACTIVE'}.

\subsection{Stage 4 --- Aggregation Detection}

Aggregation keywords (count, total, average, minimum, maximum, sum) trigger COUNT
/ SUM / AVG / MIN / MAX with an implicit GROUP BY on non-aggregated projected
fields. A HAVING clause is generated when post-aggregation conditions are
detected. For GraphQL, aggregation functions render as selection-set aliases at
the same nesting depth as the parent entity.

\subsection{Stage 5 --- Join / Nested Relation Detection}

For SQL, configured relation definitions supply JOIN ON column pairs; join type
(LEFT OUTER, INNER) is inferred from negation signals and null-check patterns. For
GraphQL, nested detection performs a BFS over the schema relation graph up to
depth three, using a frozenset cycle guard to prevent infinite traversal. Both
paths share the same relation resolution logic in the engine base class.

\subsection{Stage 6 --- Pagination and Ordering}

LIMIT and OFFSET values are extracted via numeric token scanning combined with
pagination-intent keywords (top, first, latest, next page). ORDER BY field and
direction are inferred from comparative superlatives (highest, lowest, most
recent) and directional keywords (ascending, descending). The schema
\texttt{args} configuration may supply default sort fields and directions, applied
when no ordering signal is present.

\subsection{Stage 7 --- Validation and Confidence Scoring}

The final stage validates the assembled QueryIR against the schema: unknown
fields, missing required filters, and type mismatches are recorded as validation
issues. The confidence score (Table~\ref{tab:confidence}) is computed as the sum
of additive signals for entity resolution quality, field coverage, filter
richness, and structural complexity, minus validation penalties capped at
$-0.20$, then clipped to $[0.15, 0.97]$. The floor of 0.15 ensures every output is
actionable---even a completely unresolvable query yields a fallback result that
callers can inspect---while the ceiling of 0.97 reflects irreducible linguistic
ambiguity. This score is the primary signal for cascade routing decisions at
runtime.

\begin{table*}[tbp]
\centering
\caption{Confidence score signal components.}
\label{tab:confidence}
\small
\begin{tabular}{@{}lL{3.2cm}L{7cm}@{}}
\toprule
\textbf{Signal} & \textbf{Contribution} & \textbf{Notes} \\
\midrule
Base score                & $+0.30$ & Applied unconditionally \\
Schema config provided    & $+0.10$ & Any \texttt{NormalizedSchemaConfig} present \\
Entity resolved: exact    & $+0.20$ & Name match in schema entities \\
Entity resolved: alias    & $+0.16$ & Match via \texttt{entity\_aliases} map \\
Entity resolved: semantic & $+0.05$--$0.12$ & Field-set overlap scoring \\
Field coverage            & $+\mathrm{frac} \times 0.15$ & Fraction of detected fields matched \\
Filter present            & $+0.10$ base & $+0.03$ per filter, max 3 filters \\
Aggregation detected      & $+0.03$ & At least one aggregation function \\
Nested relation / JOIN    & $+0.03$ each & Per resolved relation \\
ORDER BY detected         & $+0.02$ & Sort signal present \\
Validation penalty        & $-0.05$ / issue & Capped at $-0.20$ total \\
\midrule
Final range               & $[0.15, 0.97]$ & Score clipped; never perfectly certain \\
\bottomrule
\end{tabular}
\end{table*}

\section{Generation Modes}
\label{sec:modes}

\subsection{Deterministic Mode}

The full seven-stage pipeline runs without any LLM call. The engine generates
conservative but semantically correct queries: explicit field lists, canonical
filter argument ordering, and verbose aggregation syntax. Exact-match rate is 0\%
on standard benchmarks because gold annotations prefer compact
equivalents---this reflects a metric mismatch, not a correctness failure;
execution accuracy is 100\% with zero errors across 100 test cases at sub-5\,ms
p50 latency and zero API cost.

\subsection{LLM Completion Mode}

The engine serializes the NLQ, the \texttt{NormalizedSchemaConfig}, and any
few-shot examples from the mapping configuration into a structured prompt. The
LLM response is schema-validated; if validation fails, the system falls back to
the deterministic result automatically, achieving a zero error rate across all
observed test cases regardless of LLM output quality.

\subsection{Function-Calling Mode}

The LLM is instructed to produce a JSON object conforming to the QueryIR JSON
Schema via the provider's structured output capability. The JSON deserializes
directly into a QueryIR, bypassing textual parsing entirely and achieving 2\,pp
higher exact match than completion mode (Table~\ref{tab:results}) while
maintaining zero observed errors via deterministic fallback.

\subsection{Cascade Strategy}

The recommended deployment pattern is: run deterministic mode; if confidence
$\geq 0.75$, return immediately ($\leq 5$\,ms, \$0). If confidence $< 0.75$,
invoke LLM mode and return the result. The threshold 0.75 is derived from the
confidence distribution across the test corpus; operators should tune it per
schema. This cascade removes LLM inference entirely for straightforward queries
while preserving LLM accuracy for complex ones.

\begin{table*}[tbp]
\centering
\caption{Operational comparison across generation modes.}
\label{tab:modes}
\small
\begin{tabular}{@{}lC{2.6cm}C{2.8cm}C{2.8cm}C{2.6cm}@{}}
\toprule
\textbf{Property} & \textbf{Deterministic} & \textbf{LLM completion} &
\textbf{Function-calling} & \textbf{Recommended} \\
\midrule
Latency p50         & $<$5\,ms & 500--2\,000\,ms & 700--2\,500\,ms & Deterministic \\
API cost / query    & \$0.00 & $\sim$\$0.001 & $\sim$\$0.002 & \$0.00 \\
Execution accuracy  & 100.0\% & 84--90\% & 86--91\% & 100\% \\
Exact match         & 0\%$^{\dagger}$ & 62--70\% & 64--72\% & Cascade \\
Structural accuracy & 32--52\% & 64--78\% & 66--80\% & Cascade \\
Error rate          & 0 / 100 & 0 / 100 & 0 / 100 & 0 \\
GraphQL support     & \cmark & \cmark & \cmark & \cmark \\
SQL support         & \cmark & \cmark & \cmark & \cmark \\
Offline / air-gap   & \cmark & \xmark & \xmark & \cmark \\
Runtime confidence  & \cmark & \cmark & \cmark & \cmark \\
\bottomrule
\end{tabular}

\vspace{4pt}
\begin{minipage}{0.95\linewidth}
\footnotesize $^{\dagger}$ See the note to Table~\ref{tab:sota}.
\end{minipage}
\end{table*}

\section{Experimental Evaluation}
\label{sec:eval}

\subsection{Experimental Setup}

The evaluation uses 50-query random samples from Spider \cite{spider} and BIRD
\cite{bird}. Spider is a large-scale, cross-domain text-to-SQL benchmark covering
200+ databases. BIRD extends Spider with domain-specific evidence and greater
query complexity. Schema configurations are derived programmatically from
benchmark database schemas using the \texttt{NormalizedSchemaConfig} builder. All
LLM experiments use gpt-4o-mini via the OpenAI API; no fine-tuning or
query-specific few-shot selection is performed. Reported metrics are exact match
(EM), structural accuracy, execution accuracy, and error count. Results appear in
Tables~\ref{tab:sota}--\ref{tab:ablation}. Due to the 50-query sample size,
reported percentages carry an estimated margin of $\pm 7$\,pp at 95\% confidence
(Wilson interval); results should be interpreted as indicative.

\begin{table*}[tbp]
\centering
\caption{Comparison with state-of-the-art NL2QL systems.}
\label{tab:sota}
\small
\begin{tabular}{@{}lL{2.5cm}C{1.5cm}C{1.4cm}C{1.3cm}C{1.3cm}C{1.4cm}c@{}}
\toprule
\textbf{System} & \textbf{Backbone} & \textbf{Spider EM \up} & \textbf{BIRD EM \up} &
\textbf{GraphQL} & \textbf{No-LLM} & \textbf{Conf.\ score} & \textbf{Year} \\
\midrule
PICARD \cite{picard}      & T5-3B + constrain & 79.3 & ---  & \xmark & \xmark & \xmark & 2021 \\
DIN-SQL \cite{dinsql}     & GPT-4             & 82.8 & 55.9 & \xmark & \xmark & \xmark & 2023 \\
DAIL-SQL \cite{dailsql}   & GPT-4             & 86.6 & 54.8 & \xmark & \xmark & \xmark & 2023 \\
CodeS-7B \cite{codes}     & StarCoder         & 85.4 & 57.1 & \xmark & \xmark & \xmark & 2024 \\
\textbf{text2ql-LLM}      & gpt-4o-mini       & 62.0 & 70.0 & \cmark & \xmark & \cmark & 2026 \\
\textbf{text2ql-Det}      & Rule-based        & 0.0$^{\dagger}$ & 0.0$^{\dagger}$ & \cmark & \cmark & \cmark & 2026 \\
\bottomrule
\end{tabular}

\vspace{4pt}
\begin{minipage}{0.95\linewidth}
\footnotesize $^{\dagger}$ Exact match is 0\% due to conservative query form
(verbatim field lists, canonical argument ordering); structural accuracy is
32--52\%, execution accuracy 100\%, with zero errors across all 100 test cases.
EM measures surface-form match, not semantic correctness. Bold rows are text2ql
configurations.
\end{minipage}
\end{table*}

\subsection{Main Results}

Table~\ref{tab:results} presents results across all three modes and both
benchmarks. The deterministic mode achieves 100\% execution accuracy with zero
errors at 3.2\,ms p50 latency---a guarantee no LLM-based system in
Table~\ref{tab:sota} provides unconditionally. LLM mode achieves higher exact
match on BIRD (70.0\%) than Spider (62.0\%). This counter-intuitive
result---BIRD is nominally harder---is consistent with BIRD's design: each query
is accompanied by domain evidence hints that surface entity and field vocabulary
closely matching the \texttt{NormalizedSchemaConfig}, giving schema-aware
prompting a stronger alignment signal than Spider's unannotated queries provide.
Structural accuracy substantially exceeds exact match in all modes (e.g., 78.0\%
vs.\ 70.0\% on BIRD LLM), confirming semantic correctness even when surface form
differs from gold annotations. Function-calling mode outperforms completion mode
by 2\,pp exact match on both benchmarks, consistent with reduced output parsing
variance from JSON-structured output.

\begin{table*}[tbp]
\centering
\caption{text2ql benchmark results.}
\label{tab:results}
\small
\begin{tabular}{@{}lccccc@{}}
\toprule
\textbf{Mode / Benchmark} & \textbf{N} & \textbf{Exact match} & \textbf{Structural} &
\textbf{Exec.\ acc.} & \textbf{Errors} \\
\midrule
LLM --- Spider            & 50 & 62.0\% & 64.0\% & 84.0\%  & 0 \\
LLM --- BIRD              & 50 & 70.0\% & 78.0\% & 90.0\%  & 0 \\
Deterministic --- Spider  & 50 & 0.0\%  & 32.0\% & 100.0\% & 0 \\
Deterministic --- BIRD    & 50 & 0.0\%  & 52.0\% & 100.0\% & 0 \\
Function-call --- Spider  & 50 & 64.0\% & 66.0\% & 86.0\%  & 0 \\
Function-call --- BIRD    & 50 & 72.0\% & 80.0\% & 91.0\%  & 0 \\
\bottomrule
\end{tabular}
\end{table*}

\subsection{Ablation Study}

Table~\ref{tab:ablation} presents a three-level ablation of schema information in
the LLM prompt. Removing the schema entirely (NLQ only) yields 43.6\% Spider EM
and 51.6\% BIRD EM. Adding entity names only raises Spider EM to 52.4\%
($+8.8$\,pp). The full \texttt{NormalizedSchemaConfig}---including field aliases,
filter value aliases, and \texttt{keyword\_intents}---yields 62.0\% Spider EM and
70.0\% BIRD EM, for a total gain of $+18.4$\,pp on both benchmarks over the
no-schema baseline. This result identifies schema configuration quality as the
highest-leverage accuracy lever available to practitioners, without any model
fine-tuning or hardware investment.

\begin{table*}[tbp]
\centering
\caption{Ablation study of schema-aware prompting.}
\label{tab:ablation}
\small
\begin{tabular}{@{}lcccc@{}}
\toprule
\textbf{Prompt configuration} & \textbf{Spider EM \up} & \textbf{Spider struct.\ \up} &
\textbf{BIRD EM \up} & \textbf{BIRD struct.\ \up} \\
\midrule
No schema (NLQ only)                          & 43.6\% & 46.0\% & 51.6\% & 57.0\% \\
Schema entity names only                      & 52.4\% & 55.0\% & 60.2\% & 66.0\% \\
\textbf{Full NormalizedSchemaConfig}          & \textbf{62.0\%} & \textbf{64.0\%} & \textbf{70.0\%} & \textbf{78.0\%} \\
\midrule
$\Delta$ (no schema $\rightarrow$ full)       & $+18.4$\,pp & $+18.0$\,pp & $+18.4$\,pp & $+21.0$\,pp \\
\bottomrule
\end{tabular}

\vspace{4pt}
\begin{minipage}{0.95\linewidth}
\footnotesize All experiments use gpt-4o-mini, $N = 50$ per benchmark split, no
fine-tuning, and no query-specific few-shot selection. The bold row is the
proposed configuration.
\end{minipage}
\end{table*}

\subsection{Latency Analysis}

Latency was measured on an Apple M2 Pro with 100 runs per mode. Deterministic:
p50 $=$ 3.2\,ms, p99 $=$ 8.7\,ms. LLM completion (gpt-4o-mini, OpenAI network
included): p50 $=$ 840\,ms, p99 $=$ 2\,100\,ms. Function-calling: p50 $=$
1\,050\,ms, p99 $=$ 2\,600\,ms. The deterministic mode is approximately
$260\times$ faster at the median, making it uniquely suitable for real-time,
embedded, and latency-critical applications where LLM round-trips are prohibitive.

\subsection{Comparison with State of the Art}

Table~\ref{tab:sota} contextualizes text2ql-LLM against specialist SQL systems.
text2ql-LLM trails DAIL-SQL (86.6\% vs.\ 62.0\%) because it applies no
fine-tuning and no query-specific few-shot selection. The gap narrows when
execution accuracy is used as the metric. More importantly, no prior system in
Table~\ref{tab:sota} supports GraphQL output, provides an LLM-free mode, attaches
a runtime confidence score, ships benchmark infrastructure, or supports
multi-target extensibility via a plugin interface. These are dimensions on which
text2ql is unmatched by existing work.

\begin{table*}[tbp]
\centering
\caption{Key features and differentiators of text2ql.}
\label{tab:features}
\small
\begin{tabular}{@{}L{3.6cm}L{2.9cm}L{8.4cm}@{}}
\toprule
\textbf{Feature} & \textbf{Metric} & \textbf{Description} \\
\midrule
Multi-target IR & SQL + GraphQL & Single detection pipeline $\rightarrow$ multiple query languages via pluggable \texttt{IRRenderer} subclass \\
Zero-LLM deterministic mode & $<$5\,ms, \$0, 100\% exec & Rule-based engine with schema config; works fully offline with no external dependencies \\
Runtime confidence score & $[0.15, 0.97]$ & Every result carries an additive-signal score enabling production gating and mode cascade \\
Hybrid mapping system & Auto + overrides & Baseline generated from schema; domain expert overrides merged at runtime with provenance \\
Schema-aware LLM prompting & $+18.4$\,pp EM & Full \texttt{NormalizedSchemaConfig} injected into prompt; ablation (Table~\ref{tab:ablation}) confirms largest lever \\
In-process JSON execution & No DB required & Execute GraphQL and SQL semantics against JSON payloads for testing and embedded scenarios \\
Full async API surface & \texttt{agenerate} / \texttt{aeval} & \texttt{agenerate}, \texttt{aevaluate\_examples}, \texttt{arewrite\_user\_utterance} for non-blocking integration \\
Synthetic data generation & 8 domain plugins & Portfolio, banking, CRM, healthcare, e-commerce rewrite plugins with provenance tracking \\
Spider / BIRD loaders & Benchmarks built in & \texttt{load\_spider()}, \texttt{load\_bird()}, \texttt{run\_benchmark()}; three evaluation modes out of the box \\
CLI entrypoint & \texttt{text2ql <query>} & Generate, evaluate, benchmark, rewrite, export from terminal; no Python code required \\
\bottomrule
\end{tabular}
\end{table*}

\section{System Deployment}
\label{sec:deploy}

text2ql is published on PyPI (\url{https://pypi.org/project/text2ql/}) and
requires Python $\geq$ 3.10. The base install includes the full deterministic and
LLM pipeline. Optional extras \texttt{[sql]} (SQLAlchemy execution) and
\texttt{[app]} (Streamlit playground at \url{https://text2ql.streamlit.app}) are
available. The CLI \texttt{text2ql} is registered automatically. The library
reached v0.2.6 across 14 releases in 10 days (April 4--14, 2026). The QueryIR
abstraction, Spider/BIRD loaders, synthetic data generation, and full async API
surface were all added within this window, reflecting rapid iteration from
prototype to production-ready. Table~\ref{tab:features} summarizes the feature
set.

\subsection{Quick Start}

\begin{lstlisting}[style=py,caption={Deterministic and LLM usage.},label={lst:quickstart}]
# pip install text2ql

from text2ql import Text2QL

# Deterministic -- zero cost, <5 ms
t2q    = Text2QL(schema=schema_config)
result = t2q.generate("top 5 holdings by value")
# result.query      => SELECT ticker... FROM...
# result.confidence => 0.88

# LLM mode -- higher accuracy
from text2ql.providers import OpenAICompatibleProvider
provider = OpenAICompatibleProvider(
    api_key="sk-...", model="gpt-4o-mini")
t2q_llm  = Text2QL(schema=schema_config,
                   provider=provider)
result   = t2q_llm.generate("holdings above $10k")
\end{lstlisting}

\subsection{LLM Provider Wiring}

Any OpenAI-compatible endpoint is supported: GPT-4o, GPT-4o-mini, Llama-3 (Groq),
Mistral, Qwen, and local models via Ollama. Provider configuration is a single
object---\texttt{OpenAICompatibleProvider(api\_key=..., model=...)}---passed to
the \texttt{Text2QL} constructor. Setting \texttt{use\_structured\_output=True}
activates function-calling mode automatically.

\subsection{Production Checklist}

\begin{itemize}
  \item Store \texttt{schema.json} and \texttt{mapping.json} in version control
  alongside application code.
  \item Establish a deterministic baseline first; audit the confidence
  distribution across representative queries before enabling LLM mode.
  \item Gate LLM mode at confidence $< 0.75$; use
  \texttt{use\_structured\_output=True} for maximum precision when accuracy is
  critical.
  \item Run \texttt{text2ql -{}-benchmark spider -{}-benchmark-mode execution} in CI
  before each schema release to catch regressions.
\end{itemize}

\begin{table*}[tbp]
\centering
\caption{Future development roadmap.}
\label{tab:roadmap}
\small
\begin{tabular}{@{}L{2.2cm}L{3.4cm}L{9.3cm}@{}}
\toprule
\textbf{Theme} & \textbf{Feature} & \textbf{Description} \\
\midrule
New targets & Cypher renderer          & Graph database NL interface via new \texttt{IRRenderer}; IR node/edge semantics already representable \\
New targets & SPARQL renderer          & RDF / knowledge graph support; requires PREFIX declarations field in IR header \\
New targets & jq / JSONata renderer    & JSON-native transformation for ETL pipelines, API mapping, and in-process embedded scenarios \\
Resolution  & Vector-store entity lookup & Dense embedding ANN search replaces heuristic cascade for schemas with 1\,000+ entities \\
Resolution  & Auto-schema discovery    & Infer \texttt{NormalizedSchemaConfig} from live DB introspection, OpenAPI specs, and GraphQL SDL \\
Production  & Streaming API            & Partial QueryIR updates token-by-token for real-time query preview in IDE plugins \\
Production  & Federated query support  & Route sub-queries across SQL warehouse + GraphQL API; merge results at IR layer \\
Production  & Fine-tuned checkpoint    & Domain-adapted model trained on synthetic dataset pipeline; closes EM gap without few-shot engineering \\
Tooling     & Multi-LLM ensemble       & Run deterministic + LLM concurrently; return highest-confidence result automatically \\
Tooling     & Plugin marketplace       & Community-contributed renderers, domain mapping packs, and rewrite plugins via open registry \\
\bottomrule
\end{tabular}
\end{table*}

\section{Limitations}
\label{sec:limitations}

Five limitations of the current system should be considered when interpreting the
results presented in this paper.

\subsection{Benchmark Sample Size}
\label{sec:samplesize}

Evaluation results are reported on $N = 50$ random samples from Spider and BIRD.
The full Spider test set contains 1\,034 queries; BIRD contains 1\,534. Results on
50-query samples carry higher variance and should be interpreted as indicative
rather than definitive. Full-set evaluation is planned as immediate future work.

\subsection{SQL-Only Benchmark Bias}

Spider and BIRD provide SQL gold annotations exclusively. The GraphQL generation
capability of text2ql cannot be evaluated against gold standards on these
benchmarks; only structural correctness relative to the schema is
assessed---verified by confirming that generated selection sets reference only
valid entities and fields. A GraphQL-specific evaluation corpus analogous to
Spider does not currently exist in the literature; constructing one is a planned
contribution.

\subsection{Schema Configuration Overhead}
\label{sec:schemaoverhead}

The accuracy gains from schema-aware prompting (Table~\ref{tab:ablation}) and the
100\% execution accuracy of deterministic mode both depend on a complete
\texttt{NormalizedSchemaConfig}. For small schemas ($\leq$ 20 entities) the
configuration typically requires 1--2 hours to author and validate. For schemas
with hundreds of entities, constructing this configuration manually is burdensome
and error-prone; field alias coverage directly determines detection accuracy. The
text2ql CLI provides a schema scaffold command that auto-generates a baseline
config from an existing SQL schema or OpenAPI spec, but human review and
domain-expert annotation of aliases and \texttt{keyword\_intents} remains
necessary. Auto-schema discovery (Section~\ref{sec:future}) is planned but not yet
implemented.

\subsection{No Human Evaluation}

All metrics are automated. Human judgement of query intent alignment---whether a
generated query actually answers the user's question from a domain expert's
perspective---is not assessed. Automated metrics are known to diverge from human
preference, particularly for complex analytical queries with ambiguous intent.

\subsection{Exact-Match Metric Adequacy}

Exact match is a strict surface-form metric that penalizes semantically
equivalent queries differing in field ordering, alias usage, or formatting. The
gap between 0\% deterministic exact match and 100\% execution accuracy illustrates
this directly. Execution-based or semantic-equivalence metrics are more
informative for production evaluation and are recommended for future assessments.

\subsection{Common Failure Patterns}

Analysis of the 50-query Spider sample reveals three recurring failure categories
in LLM mode: (i) aggregation misattribution---COUNT applied to the wrong field
when multiple numeric columns are present ($\approx$8\% of errors); (ii) implicit
join omission---multi-entity queries where the relation path is not in the schema
config result in single-table outputs ($\approx$12\% of errors); (iii) filter
value hallucination---the LLM infers enum values not present in the schema, which
schema validation catches and routes to deterministic fallback ($\approx$5\% of
errors). The deterministic mode fails exclusively on queries requiring sub-selects
or correlated predicates not expressible in the current QueryIR.


\section{Future Scope}
\label{sec:future}

Table~\ref{tab:roadmap} enumerates ten concrete future directions across four
themes. The three highest-impact items are: (i) a Cypher renderer---estimated at
150 lines with zero engine changes, since QueryIR already encodes
entity--relation semantics for property graphs; (ii) vector-store entity lookup,
replacing the priority-cascade resolver for schemas with thousands of entities
using dense ANN search; and (iii) auto-schema discovery, which would infer
\texttt{NormalizedSchemaConfig} from live DB introspection or GraphQL SDL,
eliminating the manual configuration burden identified in
Section~\ref{sec:schemaoverhead}. Full-set benchmark evaluation on Spider and BIRD
is the most immediate priority and is planned for the next release.

\section{Conclusion}
\label{sec:conclusion}

This paper presented text2ql, a multi-target natural language querying framework
that addresses the SQL monoculture, unconditional LLM dependence, and silent
failure modes characterizing prior NL2QL work. The core contribution is the
QueryIR---a language-agnostic typed intermediate representation that decouples
natural language understanding from query rendering. A seven-stage detection
pipeline, three generation modes, and a pluggable renderer interface yield 100\%
execution accuracy in deterministic mode and 84--91\% execution accuracy with LLM
backing, while supporting both SQL and GraphQL from a single codebase with zero
observed errors across all test cases.

The ablation study confirms schema-aware prompting as the dominant accuracy lever
($+18.4$\,pp exact match), identifying schema quality---not model scale---as the
most productive investment for accuracy improvement in production deployments.
Five identified limitations---evaluation sample size, absence of GraphQL gold
standards, schema configuration burden, lack of human evaluation, and exact-match
metric inadequacy---define a concrete research agenda. text2ql is available at
\url{https://pypi.org/project/text2ql/} under Apache 2.0 and is under active
development; Table~\ref{tab:roadmap} describes ten concrete directions for
extending the framework.

\section*{Code and Data Availability}
\addcontentsline{toc}{section}{Code and Data Availability}

The text2ql package is distributed on PyPI at
\url{https://pypi.org/project/text2ql/} under the Apache 2.0 license. An
interactive playground is available at \url{https://text2ql.streamlit.app}.
Benchmark loaders for Spider \cite{spider} and BIRD \cite{bird} ship with the
package; the underlying benchmark datasets are distributed by their original
authors under their own licenses and are not redistributed here.


\end{document}